\documentclass[11pt,a4paper]{article}
\usepackage[hyperref]{emnlp2020}
\usepackage{times}
\usepackage{latexsym}

\usepackage{microtype}
\usepackage{graphicx}  
\usepackage{booktabs}
\usepackage{times}
\usepackage{latexsym}
\usepackage{multirow}
\usepackage{textcomp}
\usepackage[makeroom]{cancel}

\usepackage{url}
\usepackage{breakurl}
\usepackage[utf8]{inputenc}
\usepackage{graphicx}
\usepackage{array}
\usepackage{amsmath}
\usepackage{bm}
\usepackage[font=singlespacing]{caption}
\usepackage{fancyhdr}
\usepackage{amssymb}
\usepackage{nccmath}
\usepackage{color}
 \usepackage[normalem]{ulem}
 \usepackage{lipsum}
 \usepackage{graphics}
 \usepackage{tikz-dependency}
 \usepackage{enumitem}
\usepackage{tikz-qtree}
\usepackage{tikz}
 \usepackage{pgfplots}
  \usepackage{pgfplotstable}
  \usepackage{relsize}
  \usepackage[normalem]{ulem}

\usepackage{url}
\usepackage{float}
\usepackage{placeins}

\aclfinalcopy 

\newcommand{\lpk}[1]{\textcolor{blue}{{\textbf{[#1 --\textsc{Lingpeng}]}}}}
\newcommand{\ask}[1]{\textcolor{orange}{{\textbf{[#1 --\textsc{Adhi}]}}}}
\newcommand{\df}[1]{\textcolor{brown}{{\textbf{[#1 --\textsc{Daniel}]}}}}
\newcommand{\lr}[1]{\textcolor{purple}{{\textbf{[#1 --\textsc{Laura}]}}}}
\newcommand{\dy}[1]{\textcolor{red}{{\textbf{[#1 --\textsc{Dani}]}}}}
\newcommand{\cjd}[1]{\textcolor{cyan}{{\textbf{[#1 --\textsc{Chris}]}}}}
\newcommand{\pb}[1]{\textcolor{green}{\textbf{[#1 --\textsc{Phil}]}}}
\newcommand{\ignore}[1]{}

\renewcommand{\lpk}[1]{}
\renewcommand{\ask}[1]{}
\renewcommand{\df}[1]{}
\renewcommand{\lr}[1]{}
\renewcommand{\dy}[1]{}
\renewcommand{\cjd}[1]{}
\renewcommand{\pb}[1]{}

\DeclareMathOperator*{\pop}{pop}
\DeclareMathOperator*{\push}{push}
\DeclareMathOperator*{\argmax}{arg\,max}
\DeclareMathOperator*{\argmin}{arg\,min}

\newenvironment{itemizesquish}{\begin{list}{\labelitemi}{\setlength{\itemsep}{0em}\setlength{\labelwidth}{0.5em}\setlength{\leftmargin}{\labelwidth}\addtolength{\leftmargin}{\labelsep}}}{\end{list}}

\title{Syntactic Structure Distillation Pretraining For Bidirectional Encoders}

\author{Adhiguna Kuncoro\thanks{$^\star$Equal contribution.}$^{\star\spadesuit\diamondsuit}$ ~ Lingpeng Kong$^{\star\spadesuit}$ ~ Daniel Fried $^{\clubsuit}$\\\textbf{Dani Yogatama}$^{\spadesuit}$ ~ \textbf{Laura Rimell}$^{\spadesuit}$ ~ \textbf{Chris Dyer}$^{\spadesuit}$  ~ \textbf{Phil Blunsom}$^{\spadesuit\diamondsuit}$ \\
$^{\spadesuit}$DeepMind, London, UK \\
$^{\diamondsuit}$Department of Computer Science, University of Oxford, UK\\
$^{\clubsuit}$Computer Science Division, University of California, Berkeley, CA, USA\\
{\small \tt \{akuncoro,lingpenk,dyogatama,laurarimell,cdyer,pblunsom\}@google.com} \\
{\small \tt dfried@cs.berkeley.edu}
}

\date{}

\begin{document}
\maketitle
\begin{abstract}
Textual representation learners trained on large amounts of data have achieved notable success on downstream tasks; intriguingly, they have also performed well on challenging tests of syntactic competence. Given this success, it remains an open question whether 
scalable learners like BERT can become fully proficient in the syntax of natural language by virtue of data scale alone, or whether they still benefit from more explicit \textbf{syntactic biases}. To answer this question, we introduce a knowledge distillation strategy for injecting syntactic biases into BERT pretraining, by distilling the syntactically informative predictions of a hierarchical---albeit harder to scale---syntactic language model. Since BERT 
models masked words in bidirectional context, we propose to distill the approximate marginal distribution over words in context from the syntactic LM. Our approach reduces relative error by 2-21\% on a diverse set of structured prediction tasks, 
although we obtain mixed results on the GLUE benchmark. Our findings demonstrate the benefits of syntactic biases, even in representation learners that exploit large amounts of data, and contribute to a better understanding of where syntactic biases are most helpful in benchmarks of natural language understanding.

\end{abstract}

\section{Introduction}

Large-scale textual representation learners trained with variants of the language modelling (LM) objective have achieved remarkable success on downstream tasks \citep{peters_2018,devlin_2019,xlnet}. Furthermore, these models have also been shown to perform remarkably well at syntactic grammaticality judgment tasks \citep{goldberg_2019}, and encode substantial amounts of syntax in their learned representations \citep{liu_2019,tenney_2019,tenney_2018,hewitt_2019,jawahar_2019}. Intriguingly, the success on these syntactic tasks has been achieved by Transformer architectures \citep{vaswani_2017} that lack explicit notions of {\it hierarchical} syntactic structures.

Based on such evidence, it would be tempting to conclude that data scale alone is all we need to learn the syntax of natural language. Nevertheless, recent findings that systematically compare the syntactic competence of models trained at varying data scales suggest that model \emph{inductive biases} are in fact more important than data scale for acquiring syntactic competence \citep{hu_2020}. Two natural questions, therefore, are the following: can representation learners that work well at scale still benefit from explicit \emph{syntactic biases}? And where exactly would such syntactic biases be helpful in different language understanding tasks? Here we work towards answering these questions by devising a new pretraining strategy that injects syntactic biases into a BERT \citep{devlin_2019} learner that works well at scale. 
We hypothesise that this approach can improve the competence of BERT on various tasks, which provides evidence for the benefits of syntactic biases in large-scale learners.

Our approach is based on the prior work of \citet{kuncoro_19}, who devised an effective knowledge distillation \citep[KD]{bucila:2006,dark_knowledge} procedure for improving the syntactic competence of scalable LMs that lack explicit syntactic biases. More concretely, their KD procedure utilised the predictions of an explicitly hierarchical (albeit hard to scale) syntactic LM, recurrent neural network grammars \citep[RNNGs]{rnng}, as a syntactically informed learning signal for a sequential LM that works well at scale. 

Our setup nevertheless presents a new challenge: here the BERT student is a denoising autoencoder that models a collection of conditionals for words in \emph{bidirectional} context, while the RNNG teacher is an autoregressive LM that predicts words in a \emph{left-to-right} fashion, i.e. $t_{\bm{\phi}}(x_i | \mathbf{x}_{<i})$. 
This mismatch crucially means that the RNNG's estimate of $t_{\bm{\phi}}(x_i | \mathbf{x}_{<i})$ may fail to take into account the right context $\mathbf{x}_{>i}$ that is accessible to the BERT student (\S\ref{sec:approach}). 
Hence, we propose an approach where the BERT student distills  
the RNNG's marginal distribution over words in context, $t_{\bm{\phi}}(x_i | \mathbf{x}_{<i}, \mathbf{x}_{>i})$. We develop an efficient yet effective approximation for this quantity, since exact inference is expensive owing to the RNNG's left-to-right parameterisation.

Our structure-distilled BERT model differs from the standard BERT only in its pretraining objective, and hence retains the scalability afforded by Transformer architectures and specialised hardwares like TPUs. Our approach also maintains complete compatibility with the standard BERT pipelines; the structure-distilled BERT models can simply be loaded as pretrained BERT weights, which can then be fine-tuned in the exact same fashion.

We hypothesise that the stronger syntactic biases from our new pretraining procedure are useful for a variety of natural language understanding (NLU) tasks that involve structured output spaces---including tasks like semantic role labelling (SRL) and coreference resolution that are not explicitly syntactic in nature. 
We thus evaluate our models on 6 diverse structured prediction tasks, including phrase-structure parsing (in-domain and out-of-domain), dependency parsing, SRL, coreference resolution, and a CCG supertagging probe, in addition to the GLUE benchmark \citep{glue}. 
On the structured prediction tasks, our structure-distilled BERT$_{\text{BASE}}$ reduces relative error by 2\% to 21\%. These gains are more pronounced in the low-resource scenario, suggesting that stronger syntactic biases help improve sample efficiency (\S\ref{sec:experiments}).

Despite the gains on the structured prediction tasks, we achieve mixed results on GLUE: our approach yields improvements on the corpus of linguistic acceptability \citep[CoLA]{cola}, and yet performs slightly worse on the rest of GLUE. These findings 
allude to a partial dissociation between model performance on GLUE, and on other more syntax-sensitive benchmarks of NLU.

Altogether, our findings: (i) showcase the benefits of syntactic biases, even for representation learners that leverage large amounts of data, (ii) help better understand where syntactic biases are most helpful, and (iii) make a case for designing approaches that not only work well at scale, but also integrate stronger notions of syntactic biases.

\section{Recurrent Neural Network Grammars}\label{sec:rnng}
Here we briefly describe the RNNG \citep{rnng} that we use as the teacher model. An RNNG is a syntactic LM that defines the joint probability of surface strings $\mathbf{x}$ and phrase-structure nonterminals $\mathbf{y}$, henceforth denoted as $t_{\bm{\phi}}(\mathbf{x}, \mathbf{y})$, through a series of structure-building actions that traverse the tree in a top-down, left-to-right fashion. Let $N$ and $\Sigma$ denote the set of phrase-structure non-terminals and word terminals, respectively. At each time step, the decision over the next action $a_t \in \{\text{NT}(n), \text{GEN}(w), \text{REDUCE} \}$, where $n \in N$ and $w \in \Sigma$, is parameterised by a stack LSTM \citep{dyer:2015} that encodes partial constituents. The choice of $a_t$ yields these transitions:
 \begin{itemizesquish}
 \item $a_t \in \{\text{NT}(n), \text{GEN}(w)\}$ would $\push$ the corresponding embeddings $\mathbf{e}_n$ or $\mathbf{e}_w$ onto the stack;
 \item $a_t=\text{REDUCE}$ would $\pop$ the top $k$ elements up to the last incomplete non-terminal, \textbf{compose} these elements with a separate bidirectional LSTM, and lastly $\push$ the composite phrase embedding $\mathbf{e}_{\text{phrase}}$ back onto the stack. The hierarchical inductive bias of RNNGs can be attributed to this \emph{composition function},\footnote{Not all syntactic LMs have hierarchical biases; \citet{choe:2016} modelled strings and phrase structures \emph{sequentially} with LSTMs. This model can be understood as a special case of RNNGs without the composition function.} which recursively combines smaller units into larger ones.
\end{itemizesquish}
\vspace{-2mm}
RNNGs attempt to maximise the probability of correct action sequences relative to each gold tree.\footnote{Unsupervised RNNGs \citep{kim_2019} exist, although they perform worse on measures of syntactic competence.}
\vspace{-5mm}
\paragraph{Extension to subwords.} Here we extend the RNNG to operate over subword units \citep{bpe} to enable compatibility with the BERT student. As each word can be split into an arbitrary-length sequence of subwords, we preprocess the phrase-structure trees to include an additional nonterminal symbol that represents a word sequence, as illustrated by the example ``(S (NP (WORD \emph{the}) (WORD \emph{d} \emph{\#\#og})) (VP (WORD \emph{ba} \emph{\#\#rk} \emph{\#\#s})))'', where tokens prefixed by ``\emph{\#\#}'' are subword units.\footnote{An alternative here is to represent each phrase as a flat sequence of subwords, although our preliminary experiments indicate that this approach yields worse perplexity.}
\vspace{-2mm}
\section{Approach}
\label{sec:approach}
We begin with a brief review of the BERT objective, before outlining our structure distillation approach.

\subsection{BERT Pretraining Objective}
The aim of BERT pretraining is to find model parameters $\hat{\bm{\theta}}_{\text{B}}$ that would maximise the probability of reconstructing parts of $\mathbf{x}=x_1,\cdots,x_k$ conditional on a corrupted version $c(\mathbf{x})=c(x_1),\cdots,c(x_k)$, where $c(\cdot)$ denotes the stochastic corruption protocol of \citet{devlin_2019} that is applied to each word $x_i \in \mathbf{x}$. Formally:
\begin{align}
    \hat{\bm{\theta}}_{\text{B}} = \argmin_{\bm{\theta}} \sum_{i \in M(\mathbf{x})} -\log p_{\bm{\theta}}(x_i | c(x_1), \cdots, c(x_k)), \label{eq:bert_original}
\end{align}
where $M(\mathbf{x}) \subseteq \{1,\cdots,k \}$ denotes the indices of \emph{masked tokens} that serve as reconstruction targets.\footnote{In practice, the corruption protocol $c(\cdot)$ and the reconstruction targets $M(\mathbf{x})$ are intertwined; $M(\mathbf{x})$ denotes the indices of tokens in $\mathbf{x}$ ($\sim 15\%$) that were altered by $c(\mathbf{x})$.} This masked LM objective is then combined with a next-sentence prediction loss that predicts whether the two segments in $\mathbf{x}$ are contiguous sequences. 

\subsection{Motivation}\label{sec:motivation}
\begin{figure}[t]
\centering
\includegraphics[scale=0.3]{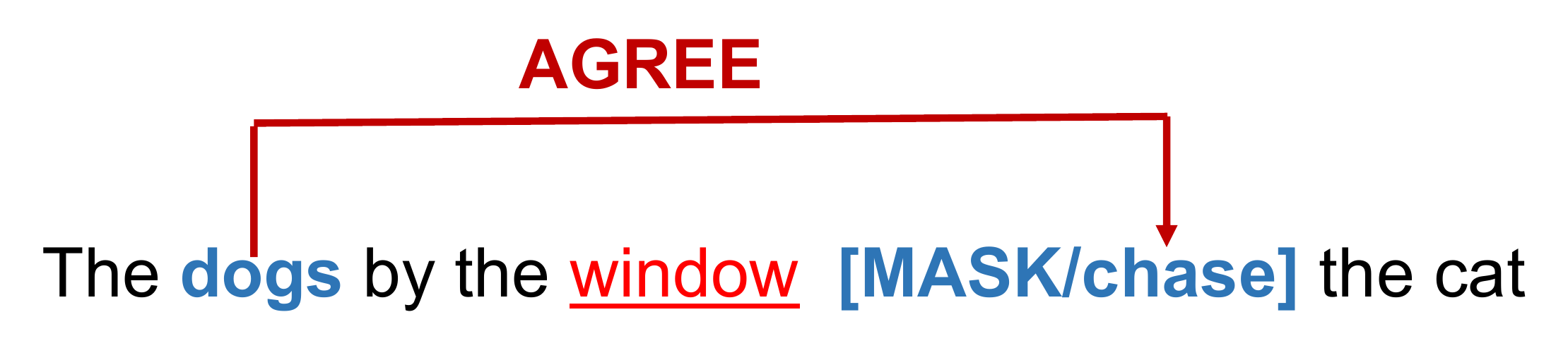}
\caption{An example of the masked LM task, where [MASK] = \emph{chase} and \emph{window} is an attractor (red). We suppress phrase-structure annotations and corruptions on the context tokens for clarity.}
\label{fig:motivation}
\end{figure}
Since the RNNG teacher is an expert on syntactic generalisations \citep{kuncoro_2018,futrell_2019,wilcox_2019}, we adopt a structure distillation procedure \citep{kuncoro_19} that enables the BERT student to learn from the RNNG's syntactically informative predictions. Our setup nevertheless means that the two models here crucially differ in nature: the BERT student is \emph{not} a left-to-right LM like the RNNG, but rather a denoising autoencoder that models a collection of conditionals for words in \textbf{bidirectional} context (Eq.~\ref{eq:bert_original}).

We now present two strategies for dealing with this challenge. The first, na\"ive approach is to \emph{ignore} this difference, and let the BERT student distill the RNNG's marginal next-word distribution for each $w \in \Sigma$ based on the left context alone, i.e. $t_{\bm{\phi}}(w| \mathbf{x}_{<i})$. While this approach is surprisingly effective (\S\ref{sec:findings}), we illustrate an issue in Fig.~\ref{fig:motivation} for ``\emph{The dogs by the window} [MASK=\emph{chase}] \emph{the cat}''. 

The RNNG's strong syntactic biases mean that we can expect $t_{\bm{\phi}}(w | \text{\emph{The dogs by the window}})$ to assign high probabilities to plural verbs like \emph{bark}, \emph{chase}, \emph{fight}, and \emph{run} that are consistent with the agreement controller \emph{dogs}---despite the presence of a singular attractor \citep{linzen-2016}, \emph{window}, that can distract the model into predicting singular verbs like \emph{chases}. Nevertheless, some plural verbs that are favoured based on the left context alone, such as \emph{bark} and \emph{run}, are in fact poor alternatives when considering the right context (e.g. ``\emph{The dogs by the window \underline{bark/run} the cat}'' are syntactically illicit). Distilling $t_{\bm{\phi}}(w | \mathbf{x}_{<i})$ thus fails to take into account the right context $\mathbf{x}_{>i}$ that is accessible to the BERT student, and runs the risk of encouraging the student to assign high probabilities for words that fit poorly with the bidirectional context. 

Hence, our second approach is to learn from teacher distributions that not only: (i) reflect the strong syntactic biases of the RNNG teacher, but also (ii) consider both the left and right context when predicting $w \in \Sigma$. Formally, we propose to distill the RNNG's marginal distribution over words in bidirectional context, $t_{\bm{\phi}}(w | \mathbf{x}_{<i}, \mathbf{x}_{>i})$, henceforth referred to as the \textbf{posterior} probability for generating $w$ under all available information. We now demonstrate that this quantity can, in fact, be computed from left-to-right LMs like RNNGs.

\subsection{Posterior Inference}\label{sec:posterior_inference}
Given a \emph{pretrained} autoregressive, left-to-right LM that factorises $t_{\bm{\phi}}(\mathbf{x}) = \prod_{i=1}^{|\mathbf{x}|} t_{\bm{\phi}}(x_i | \mathbf{x}_{<i})$, we discuss how to infer an estimate of $t_{\bm{\phi}}(x_i | \mathbf{x}_{<i}, \mathbf{x}_{>i})$. By definition of conditional probabilities:\footnote{In this setup, we assume that $\mathbf{x}$ is a fixed-length sequence, and we aim to infer the LM's estimate for generating a single token $x_i$ conditional on the full bidirectional context.}
\begin{align}
    & t_{\bm{\phi}}(x_i | \mathbf{x}_{<i}, \mathbf{x}_{>i}) = \dfrac{t_{\bm{\phi}}( \mathbf{x}_{<i}, x_i ,\mathbf{x}_{>i})}{\sum_{w \in \Sigma} t_{\bm{\phi}}( \mathbf{x}_{<i}, \tilde{x}_{i}=w ,\mathbf{x}_{>i})}, \nonumber \\
    & = \dfrac{t_{\bm{\phi}}(\mathbf{x}_{<i}) \, t_{\bm{\phi}}(x_i | \mathbf{x}_{<i})\, t_{\bm{\phi}}(\mathbf{x}_{>i} | x_i, \mathbf{x}_{<i})}{t_{\bm{\phi}}(\mathbf{x}_{<i}) \sum_{w \in \Sigma} \, t_{\bm{\phi}}(w | \mathbf{x}_{<i}) \, t_{\bm{\phi}}(\mathbf{x}_{>i} | \tilde{x_i}=w, \mathbf{x}_{<i})}, \label{eq:full_posterior} \\
     &= \dfrac{t_{\bm{\phi}}(x_i | \mathbf{x}_{<i}) \, \prod_{j=i+1}^{k} t_{\bm{\phi}}(x_j | \mathbf{x}_{<j})}{\sum_{w \in \Sigma} t_{\bm{\phi}}(w | \mathbf{x}_{<i}) \, \prod_{j=i+1}^{k} t_{\bm{\phi}}(x_j | \mathbf{\tilde{x}}_{<j}(w, i))}, \nonumber
\end{align}
where $\mathbf{\tilde{x}}_{<j}(w, i)=\left[\mathbf{x}_{<i}; w; \mathbf{x}_{i+1:j-1} \right]$ is an alternate left context where $x_i$ is replaced by $w \in \Sigma$. 
\vspace{-5mm}
\paragraph{Intuition.} After cancelling common factors $t_{\mathbf{\phi}}(\mathbf{x}_{<i})$, the posterior computation in Eq.~\ref{eq:full_posterior} is decomposed into two terms: (i) the likelihood of producing $x_i$ given its prefix, and (ii) conditional on the fact that we have generated $x_i$ and its prefix $\mathbf{x}_{<i}$, the likelihood of producing the observed continuations $\mathbf{x}_{>i}$. In our running example (Fig.~\ref{fig:motivation}), the posterior would assign low probabilities to plural verbs like \emph{bark} that are nevertheless probable under the left context alone (i.e. $t_{\bm{\phi}}($\emph{bark}$\, | \, $\emph{The dogs by the window}$)$ would be high), because they are unlikely to generate the continuations $\mathbf{x}_{>i}$ (i.e. we expect $t_{\bm{\phi}}($\emph{the cat}$\, | \,$ \emph{The dogs by the window bark}$)$ to be low since it is syntactically illicit). In contrast, the posterior would assign high probabilities to plural verbs like \emph{fight} and \emph{chase} that are consistent with the bidirectional context, since we expect both $t_{\bm{\phi}}($\emph{fight}$\, | \,$ \emph{The dogs by the window}$)$ and $t_{\bm{\phi}}($\emph{the cat}$\, | \,$\emph{The dogs by the window fight}$)$ to be probable.
\paragraph{Computational cost.} Let $k$ denote the maximum length of $\mathbf{x}$. Our KD approach requires computing the posterior distribution (Eq.~\ref{eq:full_posterior}) for every masked token $x_i$ in the dataset $D$, which (excluding marginalisation cost over $\mathbf{y}$) necessitates $O(|\Sigma| * k * |D|)$ operations, where each operation returns the RNNG's estimate of $t_{\bm{\phi}}(x_j | \mathbf{x}_{<j})$. In the standard BERT setup,\footnote{In our BERT pretraining setup, $|\Sigma| \approx 29,000$ (vocabulary size of BERT-cased), $|D| \approx 3 * 10^9$, and $k=512$.} this procedure leads to a prohibitive number of operations ($\sim5 * 10^{+16}$).
\subsection{Posterior Approximation}\label{sec:posterior_approx}
Since exact inference of the posterior is computationally expensive, here we propose an efficient approximation procedure. Approximating $t_{\bm{\phi}}(\mathbf{x}_{>i} | x_i, \mathbf{x}_{<i}) \approx t_{\bm{\phi}}(\mathbf{x}_{>i} | x_i)$ in Eq.~\ref{eq:full_posterior} yields:\footnote{This approximation preserves the intuition explained in \S\ref{sec:posterior_inference}. Concretely, verbs like \emph{bark} would also be assigned low probabilities under this approximation, since $t_{\bm{\phi}}($\emph{the cat}$\, | \,$ \emph{bark}$)$ would be low since it is syntactically illicit---the alternative ``\emph{bark \underline{\textbf{at}} the cat}'' would be syntactically licit.}
\begin{align}
   & t_{\bm{\phi}}(x_i | \mathbf{x}_{<i}, \mathbf{x}_{>i}) \approx \dfrac{t_{\bm{\phi}}(x_i | \mathbf{x}_{<i})\, t_{\bm{\phi}}(\mathbf{x}_{>i} | x_i)}{\sum_{w \in \Sigma} \, t_{\bm{\phi}}(w | \mathbf{x}_{<i}) \, t_{\bm{\phi}}(\mathbf{x}_{>i} | w)}. \label{eq:posterior_approx_1}
\end{align}
While Eq.~\ref{eq:posterior_approx_1} is still expensive to compute, it enables us to apply the Bayes rule to compute $t_{\bm{\phi}}(\mathbf{x}_{>i} | x_i)$:
\begin{align}
t_{\bm{\phi}}(\mathbf{x}_{>i} | x_i) = \dfrac{t_{\bm{\phi}}(x_i | \mathbf{x}_{>i})\, t_{\bm{\phi}}(\mathbf{x}_{>i})}{q(x_i)}, \label{eq:posterior_approx_1_5}
\end{align} 
where $q(\cdot)$ denotes the unigram distribution. For efficiency, we replace $t_{\bm{\phi}}(x_i | \mathbf{x}_{>i})$ with a separately trained ``reverse'' RNNG that operates in a \emph{right-to-left} fashion, denoted as $r_{\bm{\omega}}(x_i | \mathbf{x}_{>i})$; a complete example of the right-to-left RNNG action sequences is provided in Appendix~\ref{sec:right-to-left}. We now apply Eq.~\ref{eq:posterior_approx_1_5} and the right-to-left parameterisation $r_{\bm{\omega}}(x_i | \mathbf{x}_{>i})$ into Eq.~\ref{eq:posterior_approx_1}, and cancel common factors $t_{\bm{\phi}}(\mathbf{x}_{>i})$:
\begin{align}
   t_{\bm{\phi}}(x_i | \mathbf{x}_{<i}, \mathbf{x}_{>i}) & \approx \nonumber \\
   & \dfrac{\dfrac{t_{\bm{\phi}}(x_i | \mathbf{x}_{<i})\, r_{\bm{\omega}}(x_i | \mathbf{x}_{>i})}{q(x_i)}}{\sum_{w \in \Sigma} \dfrac{t_{\bm{\phi}}(w | \mathbf{x}_{<i})\, r_{\bm{\omega}}(w | \mathbf{x}_{>i})}{q(w)}}. \label{eq:posterior_approx_2}
\end{align}
Our approximation in Eq.~\ref{eq:posterior_approx_2} crucially reduces the required number of operations from $O(|\Sigma| * k * |D|)$ to $O(|\Sigma| * |D|)$, although the actual speedup is much more substantial in practice, since Eq.~\ref{eq:posterior_approx_2} involves easily batched operations that considerably benefit from specialised hardwares like GPUs. 

Notably, our proposed approach here is a general one; it can approximate the posterior over $x_i$ from \emph{any} left-to-right LM, which can be used as a learning signal for BERT through KD, irrespective of the LM's parameterisation. It does, however, necessitate a separately trained right-to-left LM.

\paragraph{Connection to product of experts.} Eq.~\ref{eq:posterior_approx_2} has a similar form to a product of experts \citep[PoE]{poe_hinton} between the left-to-right and right-to-left RNNGs' next-word distributions, albeit with extra unigram terms $q(w)$. If we replace the unigram distribution with a uniform one, i.e. $q(w) = 1/|\Sigma|\,\, \forall w \in \Sigma$, Eq.~\ref{eq:posterior_approx_2} reduces to a standard PoE.

\paragraph{Approximating the marginal.} The approximation in Eq.~\ref{eq:posterior_approx_2} requires estimates of $t_{\bm{\phi}}(x_i | \mathbf{x}_{<i})$ and $r_{\bm{\omega}}(x_i | \mathbf{x}_{>i})$ from the left-to-right and right-to-left RNNGs, respectively, which necessitate expensive marginalisations over all possible tree prefixes $\mathbf{y}_{<i}$ and $\mathbf{y}_{>i}$. Following \citet{kuncoro_19}, we approximate this marginalisation using a one-best predicted tree $\mathbf{\hat{y}(\mathbf{x})}=\argmax_{\mathbf{y} \in Y(\mathbf{x})} s_{\bm{\psi}}(\mathbf{y} | \mathbf{x})$, where $s_{\bm{\psi}}(\mathbf{y} | \mathbf{x})$ is parameterised by the transition-based parser of \citet{fried_2019}, and $Y(\mathbf{x})$ denotes the set of all possible trees for $\mathbf{x}$. Formally:
\begin{align}
    t_{\bm{\phi}}(x_i | \mathbf{x}_{<i}) \approx t_{\bm{\phi}}(x_i | \mathbf{x}_{<i}, \mathbf{\hat{y}}_{<i}(\mathbf{x})), \label{eq:marginal_approx}
\end{align}
where $\mathbf{\hat{y}}_{<i}(\mathbf{x})$ denotes the non-terminal symbols in $\mathbf{\hat{y}}(\mathbf{x})$ that occur before $x_i$.\footnote{Our approximation of $t_{\bm{\phi}}(x_i | \mathbf{x}_{<i})$ relies on a tree prefix $\mathbf{\hat{y}}_{<i}(\mathbf{x})$ from a separate discriminative parser, which has access to yet unseen words $\mathbf{x}_{>i}$. This non-incremental procedure is justified, however, since we aim to design the most informative teacher distributions for the non-incremental BERT student, which also has access to bidirectional context.} The marginal next-word distributions $r_{\bm{\omega}}(x_i | \mathbf{x}_{>i})$ from the right-to-left RNNG is approximated similarly.

\paragraph{Preliminary Experiments.} Before proceeding with the KD experiments, we assess the quality and feasibility of our approximation through preliminary language modelling experiments on the Penn Treebank \citep[PTB]{marcus:1993}; full details are provided in Appendix~\ref{sec:prelim}. We find that our approximation is much faster than exact inference by a factor of more than 50,000, at the expense of a slightly worse average posterior negative log-likelihood (2.68 rather than 2.5 for exact inference). 

\subsection{Objective Function}
In our structure distillation pretraining, we aim to find BERT parameters $\hat{\bm{\theta}}_{\text{KD}}$ that emulate our approximation of $t_{\bm{\phi}}(x_i | \mathbf{x}_{<i}, \mathbf{x}_{>i})$ through a word-level cross-entropy loss \citep[\emph{inter alia}]{dark_knowledge,seq-distillation,furlanello_18}:
\begin{align*}
&\hat{\bm{\theta}}_{\text{KD}} = \argmin_{\bm{\theta}} \dfrac{1}{|D|} \sum_{\mathbf{x} \in D} \ell_{\text{KD}}(\mathbf{x}; \bm{\theta}), \text{where} \\
& \ell_{\text{KD}}(\mathbf{x}; \bm{\theta}) = - \sum_{i \in M(\mathbf{x})} \sum_{w \in \Sigma} \Big[ \tilde{t}_{\bm{\phi},\bm{\omega}}(w | \mathbf{x}_{<i}, \mathbf{x}_{>i}) \\
&   \log p_{\bm{\theta}}\left(\tilde{x}_i=w | c(x_1), \cdots, c(x_k)\right) \Big],
\end{align*}
where $\tilde{t}_{\bm{\phi},\bm{\omega}}(w | \mathbf{x}_{<i}, \mathbf{x}_{>i})$ is our approximation of $t_{\bm{\phi}}(w | \mathbf{x}_{<i}, \mathbf{x}_{>i})$, as defined in Eqs.~\ref{eq:posterior_approx_2} and \ref{eq:marginal_approx}.

\paragraph{Interpolation.} The RNNG teacher is an expert on syntax, although in practice it is only feasible to train it on a much smaller dataset. Hence, we not only want the BERT student to learn from the RNNG's syntactic expertise, but also from the rich common-sense and semantics knowledge contained in large text corpora by virtue of predicting the true identity of the masked token $x_i$,\footnote{The KD loss $\ell_{\text{KD}}(\mathbf{x}; \bm{\theta})$ is defined independently of $x_i$.} as done in the standard BERT setup. We thus interpolate the KD loss and the original BERT masked LM objective:
\begin{align}
\hat{\bm{\theta}}_{\text{B-KD}} &= \argmin_{\bm{\theta}} \dfrac{1}{|D|} \sum_{\mathbf{x} \in D} \Big[ \alpha \ell_{\text{KD}}(\mathbf{x}; \bm{\theta}) + (1 - \alpha) \nonumber  \\
& \sum_{i \in M(\mathbf{x})} - \log p_{\bm{\theta}}(x_i | c(x_1), \cdots, c(x_k)) \Big], \label{eq:interpolation}
\end{align}
omitting the next-sentence prediction for brevity. We henceforth set $\alpha=0.5$ unless stated otherwise.

\section{Experiments}\label{sec:experiments}
Here we outline the evaluation setup, present our results, and discuss the implications of our findings.

\subsection{Evaluation Tasks and Setup}\label{sec:tasks}
We conjecture that the improved syntactic competence from our approach would benefit a broad range of tasks that involve structured output spaces, including those that are not explicitly syntactic. We thus evaluate our structure-distilled BERTs on six diverse structured prediction tasks that encompass syntactic, semantic, and coreference resolution tasks, in addition to the GLUE benchmark that is largely comprised of classification tasks.
\vspace{-2mm}
\paragraph{Phrase-structure parsing - PTB.} We first evaluate our model on phrase-structure parsing on the WSJ section of the PTB. Following prior work, we use sections 02-21 for training, section 22 for validation, and section 23 for testing. We apply our approach on top of the BERT-augmented in-order \citep{in_order_rnng} transition-based parser of \citet{fried_2019}, which approaches the current state of the art. Since the RNNG teacher that we distill into BERT also employs phrase-structure trees, this setup is related to self-training \citep[\emph{inter alia}]{yarowsky_1995,Charniak:1997,tri_training,mcclosky:2006,andor-etal-2016-globally}.
\vspace{-6mm}
\paragraph{Phrase-structure parsing - OOD.} Still in the context of phrase-structure parsing, we evaluate how well our approach generalises to three out-of-domain (OOD) treebanks: Brown \citep{brown_corpus}, Genia \citep{genia}, and the English Web Treebank \citep{ewt}. Following \citet{fried_2019}, we test the PTB-trained parser on the test splits\footnote{We use the Brown test split of \citet{gildea-2001-corpus}, the Genia test split of \citet{mcclosky-etal-2008-self}, and the EWT test split from SANCL 2012 \citep{ewt}.} of these OOD treebanks \emph{without} any retraining, to simulate the case where no in-domain labelled data are available. We use the same codebase as above.
\vspace{-2mm}
\paragraph{Dependency parsing - PTB.} Our third task is PTB dependency parsing with Stanford Dependencies \citep{stanford_dependencies} v3.3.0. We use the BERT-augmented joint phrase-structure and dependency parser of \citet{hpsg_2019}, which is inspired by head-driven phrase-structure grammar \citep[HPSG]{hpsg_book}.
\vspace{-2mm}
\paragraph{Semantic role labelling.} Our fourth evaluation task is span-based semantic role labelling (SRL) on the CoNLL 2012 dataset \citep{srl_conll_2012}. We apply our approach on top of the BERT-augmented model of \citet{shi_lin_2019}, as implemented in AllenNLP \citep{allennlp}.
\vspace{-2mm}
\paragraph{Coreference resolution.} Our fifth evaluation task is coreference resolution on the OntoNotes benchmark \citep{pradhan2012coref}. 
For this task, we use the BERT-augmented model of \citet{joshi2019coref}, which extends the higher-order coarse-to-fine model of \citet{lee2018c2f}.
\paragraph{CCG Supertagging Probe.} All proposed tasks thus far necessitate either fine-tuning the entire BERT model, or training a task-specific model on top of the BERT embeddings. Hence, it remains unclear how much of the gains are due to better structural representations from our new \emph{pretraining} strategy, rather than the available supervision at the \emph{fine-tuning} stage. To better understand the gains from our approach, we evaluate on combinatory categorial grammar \citep[CCG]{steedman_2000} supertagging \citep{bangalore-joshi-1999-supertagging,clark-curran-2007-wide} through a \textbf{classifier probe} \citep[\emph{inter alia}]{shi_16,adi_2017,belinkov_2017a}, where no BERT fine-tuning takes place.\footnote{A similar CCG probe was explored by \citet{liu_2019}; we obtain comparable numbers for the no distillation baseline.}

CCG supertagging is a compelling probing task since it necessitates an understanding of bidirectional context information; the per-word classification setup also lends itself well to classifier probes. Nevertheless, it remains unclear how much of the accuracy can be attributed to the information encoded in the representation, as opposed to the classifier probe itself. We thus adopt the \textbf{control task} protocol of \citet{hewitt_liang_2019} that assigns each word type to a random control category,\footnote{Following \citet{hewitt_liang_2019}, the cardinality of this control category is the same as the number of supertags.} which assesses the memorisation capacity of the classifier. In addition to the probing accuracy, we report the probe \emph{selectivity},\footnote{A probe's selectivity is defined as the difference between the probing task accuracy and the control task accurary.} where higher selectivity denotes probes that faithfully rely on the linguistic knowledge encoded in the representation. We use linear classifiers to maintain high selectivities.
\paragraph{Commonality.} All our structured prediction experiments are conducted on top of publicly available repositories of BERT-augmented models, with the exception of CCG supertagging that we evaluate as a probe. This setup means that obtaining our results is as simple as changing the pretrained BERT weights to our structure-distilled BERT, and applying the exact same steps as in the baseline. 
\vspace{-2mm}
\paragraph{GLUE.} Beyond the 6 structured prediction tasks above, we evaluate our approach on the classification\footnote{This setup excludes the semantic textual similarity benchmark (STS-B), which is formulated as a regression task.} tasks of the GLUE benchmark except the Winograd NLI \citep{winograd} for consistency with the original BERT \citep{devlin_2019}. For each GLUE task fine-tuning, we run a grid search over five potential learning rates, two batch sizes, and five random seeds (Appendix~\ref{sec:bert_params}), leading to 50 fine-tuning configurations that we run and evaluate on the validation set of each GLUE task.

\subsection{Experimental Setup and Baselines} \label{sec:experimental_setup}
Here we describe the key aspects of our empirical setup, and outline the baselines for assessing the efficacy of our approach.
\paragraph{RNNG Teacher.} We implement the subword-augmented RNNG teachers (\S\ref{sec:rnng}) on DyNet \citep{dynet}, and obtain ``silver-grade'' phrase-structure annotations for the entire BERT training set using the transition-based parser of \citet{fried_2019}. These trees are used to train the RNNG (\S\ref{sec:rnng}), and to approximate its marginal next-word distribution at inference (Eq.~\ref{eq:marginal_approx}). We use the same WordPiece tokenisation and vocabulary as BERT-Cased; Appendix \ref{sec:rnng_params} summarises the complete list of RNNG hyper-parameters. Since our approximation (Eq.~\ref{eq:posterior_approx_2}) makes use of a right-to-left RNNG, we train this variant (Appendix~\ref{sec:right-to-left}) with the same hyper-parameters and data as the left-to-right model. We train each directional RNNG teacher on a shared subset of 3.6M sentences ($\sim$3\%) from the BERT training set with automatic batching \citep{neubig_2017}, which takes three weeks on a V100 GPU.  
\paragraph{BERT Student.} We apply our structure distillation pretraining protocol to BERT$_{\text{BASE}}$-Cased,\footnote{We use BERT$_{\text{BASE}}$ rather than BERT$_{\text{LARGE}}$ to reduce the turnaround of our experiments, although our approach can easily be extended to BERT$_{\text{LARGE}}$.} using the exact same training dataset, model configuration, WordPiece tokenisation, vocabulary, and hyper-parameters (Appendix~\ref{sec:bert_params}) as in the standard pretrained BERT model.\footnote{\url{https://github.com/google-research/bert}.} The sole exception is that we use a larger initial learning rate of $3e^{-4}$ based on preliminary experiments,\footnote{We find this larger learning to perform better on most of our evaluation tasks. \citet{roberta} has similarly found that tuning BERT's initial learning rate leads to better results.} which we apply to all models (including the no distillation/standard BERT baseline) for fair comparison.
\vspace{-5mm}
\paragraph{Baselines and comparisons.} We compare the following set of models in our experiments:
\vspace{-2mm}
\begin{itemizesquish}
\item A standard BERT$_{\text{BASE}}$-Cased without any structure distillation loss, which benefits from scalability but lacks syntactic biases (\textbf{``No-KD''});
\item Four variants of structure-distilled BERTs that: (i) only distill the left-to-right RNNG (\textbf{``L2R-KD''}), (ii) only distill the right-to-left RNNG (\textbf{``R2L-KD''}), (iii) distill the RNNG's approximated marginal for generating $x_i$ under the bidirectional context, where $q(w)$ (Eq.~\ref{eq:posterior_approx_2}) is the \emph{uniform} distribution (\textbf{``UF-KD''}), and lastly (iv) a similar variant as (iii), but where $q(w)$ is the \emph{unigram} distribution (\textbf{``UG-KD''}). All these BERT models crucially benefit from the syntactic biases of RNNGs, although only variants (iii) and (iv) learn from teacher distributions that consider \emph{bidirectional context} for predicting $x_i$; and
\item A BERT$_{\text{BASE}}$ model that distills the approximated marginal for generating $x_i$ under the bidirectional context, but from \emph{sequential} LSTM teachers (\textbf{``Seq-KD''}) in place of RNNGs.\footnote{For fair comparison, we train the LSTM on the exact same subset as the RNNG, with comparable number of model parameters. An alternative here is to use Transformers, although we elect to use LSTMs to facilitate fair comparison with RNNGs, which are also based on LSTM architectures.} This baseline crucially isolates the importance of learning from hierarchical teachers, since it employs the exact same approximation technique and KD loss as the structure-distilled BERTs.
\end{itemizesquish}
\paragraph{Learning curves.} Given enough labelled data, BERT can acquire the relevant structural information from the fine-tuning (as opposed to pretraining) procedure, although better pretrained representations can nevertheless facilitate \emph{sample-efficient} generalisations \citep{tgli}. We thus additionally examine the models' fine-tuning learning curves, as a function of varying amounts of training data, on phrase-structure parsing and SRL.
\paragraph{Random seeds.} Since fine-tuning the same pretrained BERT with different random seeds can lead to varying results, we report the mean performance from three random seeds on the structured prediction tasks, and from five random seeds on GLUE.
\paragraph{Test results.} To preserve the integrity of the test sets, we first report all performance on the validation set, and only report test set results for: (i) the \textbf{No-KD} baseline, and (ii) the best structure-distilled model on the validation set (\textbf{``Best-KD''}).

\subsection{Findings and Discussion}\label{sec:findings}
\begin{figure*}[t]
\centering
\includegraphics[scale=0.6]{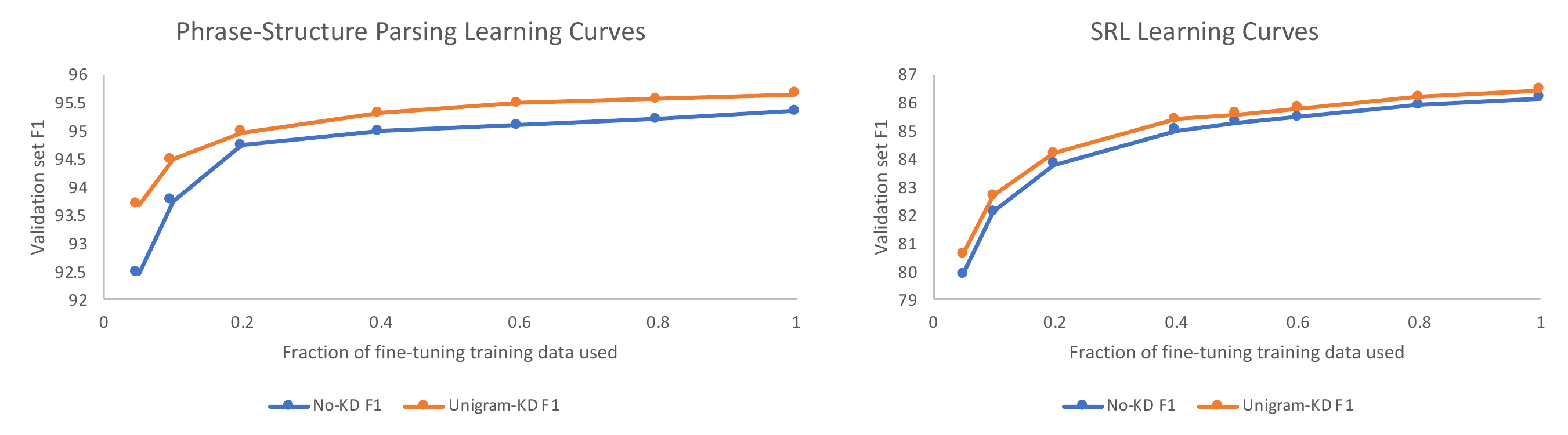}
\caption{The fine-tuning learning curves that examine how the number of fine-tuning instances (from 5\% to 100\% of the full training sets) affect validation set F1 in the case of phrase-structure parsing and SRL. We compare the \textbf{``No-KD''}/standard BERT$_{\text{BASE}}$-Cased and the \textbf{``UG-KD''} structure-distilled BERT.}
\label{fig:learning_curves}
\end{figure*}

We report the validation and test results of the structured prediction tasks in Table~\ref{tab:results_structured}. The validation set learning curves for phrase-structure parsing and SRL that compare the \textbf{No-KD} baseline and the \textbf{UG-KD} variant are provided in Fig.~\ref{fig:learning_curves}.

\begin{table*}[t]
  \centering
      \resizebox{\textwidth}{!}{%
\begin{tabular}{l|r||r|r||r|r|r|r||r|r|r} 
\toprule
\multicolumn{2}{c||}{\multirow{3}{*}{\textbf{Task}}} & \multicolumn{6}{c}{\textbf{Validation Set}} & \multicolumn{3}{||c}{\textbf{Test Set}} \\ \cline{3-11}
\multicolumn{2}{c||}{} & \multicolumn{2}{c||}{\textbf{Baselines}} &
\multicolumn{4}{c}{\textbf{Structure-distilled BERTs}}&
\multicolumn{1}{||c}{\multirow{2}{*}{\textbf{No-KD}}} &
\multicolumn{1}{|c|}{\multirow{2}{*}{\textbf{Best-KD}}} &
\multicolumn{1}{c}{\multirow{2}{*}{\textbf{Err. Red.}}} \\ \cline{3-8}
\multicolumn{2}{c||}{} & \multicolumn{1}{c|}{\textbf{No-KD}} & \multicolumn{1}{c||}{\textbf{Seq-KD}} & \multicolumn{1}{c|}{\textbf{L2R-KD}} & \multicolumn{1}{c|}{\textbf{R2L-KD}} & \multicolumn{1}{c|}{\textbf{UF-KD}} & \multicolumn{1}{c}{\textbf{UG-KD}} & \multicolumn{1}{||c}{} & \multicolumn{1}{|c|}{}  & \multicolumn{1}{c}{} \\ \hline                                                                                                                                                                                                   
\multirow{5}{*}{\rotatebox[origin=c]{90}{Parsing}} & Const. PTB - F1   & 95.38   &   95.33                        & 95.55                               & 95.55                               & 95.58                                   & \textbf{95.59}                       & 95.35  & \textbf{95.70} & 7.6\%               \\
 &Const. PTB - EM  & 55.33  &   55.41                         & 55.92                               & 56.18                               & 56.39                                  &\textbf{56.59} &        55.25                   & \textbf{57.77} & 5.63\%                                      \\
 &Const. OOD - F1$^{\dagger}$     & 87.71    &    87.23                      & 88.36                               & \textbf{88.56}                      & 88.24                                   & 88.21                                  & 89.04 & \textbf{89.76} & 6.55\%                                      \\
& Dep. PTB - UAS     & 96.48    &  96.40                        & \textbf{96.70}                      & 96.64                               & 96.60                                   & 96.66                                   & 96.79 & \textbf{96.86} & 2.18\%                                      \\
& Dep. PTB - LAS & 94.65     &    94.56                     & \textbf{94.90}                      & 94.80                               & 94.79                                   & 94.83                                   & 95.13 & \textbf{95.23} & 1.99\%                                      \\ \hline
\multicolumn{2}{l||}{SRL - CoNLL 2012}                    & 86.17   &    86.09                       & 86.34                               & 86.29                               & 86.30                                   & \textbf{86.46}                       &  86.08 & \textbf{86.39} & 2.23\%                                       \\
\multicolumn{2}{l||}{Coref.}                 & 72.53      &   69.27                     & 73.74                               & 73.49                               & \textbf{73.79}                          & 73.33                                 &  72.71 & \textbf{73.69} & 3.58\%                                      \\ \hline \hline
\multicolumn{2}{l||}{CCG supertag. probe}             & 93.69     &  91.59                       & 93.97                               & \textbf{95.21}                       & 95.13                                   & \textbf{95.21}                        &  93.88 & \textbf{95.2} & 21.57\%                                     \\
\multicolumn{2}{l||}{Probe selectivity}             & 24.79     & 23.77                        & 23.3                               & 23.57                       & 27.28                                   & \textbf{28.3}                        & 23.15  & \textbf{26.07} & N/A                                     \\
\bottomrule                                   
\end{tabular}}
\caption{Validation and test results for the structured prediction tasks; each entry reflects the mean of three random seeds. To preserve test set integrity, we only obtain test set results for the no distillation baseline and the best structure-distilled BERT on the validation set; \textbf{``Err. Red.''} reports the test error reductions \emph{relative} to the \textbf{No-KD} baseline. We report F1 and exact match (EM) for PTB phrase-structure parsing; for dependency, we report unlabelled (UAS) and labelled (LAS) attachment scores. The ``Const. OOD'' ($^{\dagger}$) row indicates the mean F1 from three out-of-domain corpora: Brown, Genia, and the English Web Treebank (EWT), although the validation results exclude the Brown Treebank that has no validation set.}\label{tab:results_structured}
\end{table*}

\paragraph{General discussion.} We summarise several key observations from Table~\ref{tab:results_structured} and Fig.~\ref{fig:learning_curves}.
\begin{itemizesquish}
\item All four structure-distilled BERT models consistently outperform the \textbf{No-KD} baseline, including the \textbf{L2R-KD} and \textbf{R2L-KD} variants that only distill the syntactic knowledge of unidirectional RNNGs. Remarkably, this pattern holds true for \emph{all six} structured prediction tasks. In contrast, we observe no such gains for the \textbf{Seq-KD} baseline, which largely performs worse than the \textbf{No-KD} model. We conclude that the gains afforded by our structure-distilled BERTs can be attributed to the hierarchical bias of the RNNG teacher.

\item We conjecture that the surprisingly strong performance of the \textbf{L2R-KD} and \textbf{R2L-KD} models, which distill the knowledge of \emph{unidirectional} RNNGs, 
can be attributed to the interpolated objective in Eq.~\ref{eq:interpolation} ($\alpha=0.5$). This interpolation means that the target distribution
assigns a probability mass of at least 0.5 to the true masked word $x_i$, which is guaranteed to be consistent with the bidirectional context. 
However, the syntactic knowledge contained in the unidirectional RNNGs' predictions can still provide a structurally informative learning signal, via the rest of the probability mass, for the BERT student.

\item While all structure-distilled variants outperform the baseline, models that distill our approximation of the RNNG's distribution for words in bidirectional context (\textbf{UF-KD} and \textbf{UG-KD}) yield the best results on four out of six tasks (PTB phrase-structure parsing, SRL, coreference resolution, and the CCG supertagging probe). This finding confirms the efficacy of our  approach.

\item We observe the largest gains for the syntactic tasks, particularly for phrase-structure parsing and CCG supertagging. However, the improvements are not at all confined to purely syntactic tasks: we reduce relative error from strong BERT baselines by 2.2\% and 3.6\% on SRL and coreference resolution, respectively. While the RNNG's syntactic biases are derived from phrase-structure grammar, the strong improvement on CCG supertagging, in addition to the smaller improvement on dependency parsing, suggests that the RNNG's syntactic biases generalise well across different syntactic formalisms.

\item We observe larger improvements in a low-resource scenario, where the model is exposed to fewer fine-tuning instances (Fig.~\ref{fig:learning_curves}), suggesting that syntactic biases are helpful for enabling more \textbf{sample-efficient} generalisations. This pattern holds for both tasks that we investigated: phrase-structure parsing (syntactic) and SRL (not explicitly syntactic). With only 5\% of the fine-tuning data, the \textbf{UG-KD} model improves F1 from 79.9 to 80.6 for SRL (a 3.5\% error reduction relative to the \textbf{No-KD} baseline, as opposed to. 2.2\% on the full data). For phrase-structure parsing, the \textbf{UG-KD} model achieves a 93.68 F1 (a 16\% relative error reduction, as opposed to 7.6\% on the full data) with only 5\% of the PTB---this performance is notably better than past state of the art parsers trained on the \emph{full} PTB c. 2017 \citep{kuncoro-2017}. 

\end{itemizesquish}
\paragraph{GLUE results and discussion.} We report the GLUE validation and test results in Table~\ref{tab:glue}.
Since we observe a different pattern of results on the Corpus of Linguistic Acceptability \citep[CoLA]{cola} than on the rest of GLUE, we henceforth report: (i) the CoLA results, (ii) the 7-task average that excludes CoLA, and (iii) the average across all 8 tasks. We select the \textbf{UG-KD} model since it achieved the best 8-task average on the GLUE validation sets;
 the full GLUE breakdown for these two models is provided in Appendix~\ref{sec:full_glue}.


\begin{table}[t]
 \centering
  \resizebox{0.45\textwidth}{!}{%
\begin{tabular}{l|r|r}
\toprule
\multicolumn{1}{c|}{} & \multicolumn{1}{c|}{\textbf{No-KD}} & \multicolumn{1}{c}{\textbf{UG-KD}} \\ \midrule
\multicolumn{3}{c}{Validation Set (Per-task average / 1-best)}                                                                  \\ \midrule
CoLA                 &   50.7 / 60.2                               & \textbf{54.3}  / \textbf{60.6}                                 \\
7-task avg. (excl. CoLA) & \textbf{85.4} / \textbf{87.8} & 84.8 / 86.9 \\
Overall 8-task avg.   & \textbf{81.1} / \textbf{84.4}  &  81.0 / 83.6                                    \\ \midrule
\multicolumn{3}{c}{Test set (Per-task 1-best on validation set)}                                                                        \\ \midrule
CoLA                 &          53.1     &     \textbf{55.3}                                    \\
7-task avg. (excl. CoLA) & \textbf{84.2} & 83.5  \\
Overall 8-task avg.       & \textbf{80.3} & 80.0  \\ \bottomrule                                       
\end{tabular}}
\caption{Summary of the validation and test set results on GLUE. The validation results are derived from the average of five random seeds for each task, which accounts for variance, and the 1-best random seed, which does not. The test results are derived from the 1-best random seed on the validation set.} 
\label{tab:glue}
\end{table}


The results on GLUE provide an interesting contrast to the consistent improvement we observed on the structured prediction tasks. More concretely, our \textbf{UG-KD} model outperforms the baseline on CoLA, but performs slightly worse on the other GLUE tasks in aggregate, leading to a slightly lower overall test set accuracy (80.0 for the \textbf{UG-KD} as opposed to 80.3 for the \textbf{No-KD} baseline). 

The improvement on the syntax-sensitive CoLA provides additional evidence---beyond the improvement on the syntactic tasks (Table~\ref{tab:results_structured})---that our approach indeed yields improved syntactic competence. We conjecture that these improvements do not transfer to the other GLUE tasks because they rely more on lexical and semantic properties, and less on syntactic competence \citep{mccoy_2019}.

We defer a more thorough investigation of how much syntactic competence is necessary for solving most of the GLUE tasks to future work, but make two remarks. First, the findings on GLUE are consistent with the hypothesis that our approach yields improved structural competence, albeit at the expense of a slightly less rich meaning representation, which we attribute to the smaller dataset used to train the RNNG teacher. Second, human-level natural language understanding includes the ability to predict structured outputs, e.g. to decipher ``\emph{who did what to whom}'' (SRL). Succeeding in these tasks necessitates inference about structured output spaces, which (unlike most of GLUE) cannot be reduced to a single classification decision. Our findings indicate a partial dissociation between model performance on these two types of tasks; hence, supplementing GLUE evaluation with some of these structured prediction tasks can offer a more holistic assessment of progress in NLU.

\begin{table*}[!htb]
\centering
\resizebox{\textwidth}{!}{%
\begin{tabular}{l|c|c} \toprule
\multicolumn{1}{c|}{\textbf{Sent.}}   & \textbf{No-KD \& L2R-KD Pred.}                   & \textbf{R2L-KD \& UG-KD Pred.}                           \\ \midrule
\emph{``Apple II owners , for example , had to \textbf{\underline{use}} their TV} & \multirow{2}{*}{\textcolor{red}{(S{[}b{]}\textbackslash{}NP)/NP}} & \multirow{2}{*}{\textcolor{blue}{((S{[}b{]}\textbackslash{}NP)/PP)/NP}} \\
\emph{sets as screens and stored data on audiocassettes''} &                                                  &                                                       \\ \bottomrule
\end{tabular}}
\caption{An example of the CCG supertag predictions for the verb ``\textbf{\underline{\emph{use}}}'' from four different BERT variants. The correct answer is ``\textcolor{blue}{((S{[}b{]}\textbackslash{}NP)/PP)/NP}'', which both the \textbf{R2L-KD} and \textbf{UG-KD} predict correctly (\textcolor{blue}{blue}). However, the \textbf{No-KD} baseline and the \textbf{L2R-KD} model produce (the same) incorrect predictions (\textcolor{red}{red}); both models fail to subcategorise the prepositional phrase ``\emph{as screens}'' as a dependent of the verb ``\textbf{\underline{\emph{use}}}''. Beyond this, all four models predict the correct supertags for all other words (not shown).
}\label{tab:error_analysis}
\end{table*}

\paragraph{CCG probe example.} The CCG supertagging probe is a particularly interesting test bed, since it clearly assesses the model's ability to use contextual information in making its predictions, \emph{without} introducing additional confounds from the BERT fine-tuning procedure. 
We thus provide a representative example of four different BERT variants' predictions on the CCG supertagging probe in Table~\ref{tab:error_analysis}, based on which we discuss two observations. First, the different models make different predictions, where the \textbf{No-KD} and \textbf{L2R-KD} models produce (coincidentally the same) incorrect predictions, while the \textbf{R2L-KD} and \textbf{UG-KD} models are able to predict the correct supertag. This finding suggests that different teacher distributions are able to impose different biases on the BERT students.\footnote{All four BERTs have access to the full \emph{bidirectional} context at test time, although some are trained to mimic the predictions of \emph{unidirectional} RNNGs (\textbf{L2R-KD} and \textbf{R2L-KD}).} 

Second, the mistakes of the \textbf{No-KD} and \textbf{L2R-KD} BERTs belong to the broader category of challenging argument-adjunct distinctions \citep{palmer-etal-2005-proposition}. Here both models fail to subcategorise for the prepositional phrase (PP) ``\emph{as screens}'', which serves as an argument of the verb ``\emph{use}'', as opposed to the noun phrase ``\emph{TV sets}''. Distinguishing between these two potential dependencies naturally requires syntactic information from the right context; hence the \textbf{R2L-KD} BERT, which is trained to emulate the predictions of an RNNG teacher that observes the right context, is able to make the correct prediction. This advantage is crucially retained by the \textbf{UG-KD} model that distills the RNNG's approximate distribution over words in bidirectional context (Eq.~\ref{eq:posterior_approx_2}), and further confirms the efficacy of our proposed approach.

\vspace{-3mm}
\subsection{Limitations} We outline two limitations to our approach. First, we assume the existence of decent-quality ``silver-grade'' phrase-structure trees to train the RNNG teacher. While this assumption holds true for English due to the existence of accurate phrase-structure parsers, this is not necessarily the case for other languages. Second, pretraining the BERT student in our na\"ive implementation is about half as fast on TPUs compared to the baseline due to I/O bottleneck. This overhead only applies at pretraining, and can be reduced through parallelisation. 

\section{Related Work}
Earlier work has proposed a few ways for introducing notions of hierarchical structures into BERT, for instance through designing structurally motivated auxiliary losses \citep{wang2019structbert}, or including syntactic information in the embedding layers that serve as inputs for the Transformer \citep{sundararaman2019syntax}. In contrast, we employ a different technique for injecting syntactic biases, which is based on the structure distillation technique of \citet{kuncoro_19}, although our work features two key differences. First, \newcite{kuncoro_19} put a sole emphasis on cases where both the teacher and student models are autoregressive, left-to-right LMs; here we extend this objective for when the student model is a representation learner that has access to bidirectional context. Second, \newcite{kuncoro_19} only evaluated their structure-distilled LMs in terms of perplexity and grammatical judgment \citep{marvin_2018}. In contrast, we evaluate our structure-distilled BERT models on 6 diverse structured prediction tasks and the GLUE benchmark. It remains an open question whether, and how much, syntactic biases are helpful for a broader range of NLU tasks beyond grammatical judgment; our work represents a step towards answering this question.

More recently, substantial progress has been made in improving the performance of BERT and the broader class of masked LMs \citep[\emph{inter alia}]{lan2019albert,roberta,2019t5,ernie}. Our structure distillation technique is orthogonal, and can be applied on top of these approaches. Lastly, our findings on the benefits of syntactic knowledge for structured prediction tasks that are not explicitly syntactic in nature, such as SRL and coreference resolution, are consistent with those of prior work \citep[\emph{inter alia}]{syntactic_scaffold,he2018syntax,strubell_2018}. 



\section{Conclusion}
Given the remarkable success of textual representation learners trained on large amounts of data, it remains an open question whether syntactic biases are still relevant these models that work well at scale. Here we present evidence to the affirmative: our structure-distilled BERT models outperform the baseline on a diverse set of 6 structured prediction tasks. We achieve this through a new pretraining strategy that enables the BERT student to learn from the predictions of an explicitly hierarchical, but much less scalable, RNNG teacher model. Since the BERT student is a bidirectional model that estimates the conditional probabilities of masked words in context, we propose to distill an efficient yet surprisingly effective approximation of the RNNG's estimate for generating each word conditional on its bidirectional context.

Our findings suggest that syntactic inductive biases are beneficial for a diverse range of structured prediction tasks, including for tasks that are not explicitly syntactic in nature. In addition, these biases are particularly helpful for improving fine-tuning sample efficiency on downstream tasks. Lastly, our findings motivate the broader question of how we can design models that integrate stronger notions of structural biases---and yet can be easily scalable at the same time---as a promising (if relatively underexplored) direction of future research.

\section*{Acknowledgments}
We would like to thank Mandar Joshi, Zhaofeng Wu, and Rui Zhang for answering questions regarding the evaluation of the model. We also thank Sebastian Ruder, John Hale, Kris Cao, and Stephen Clark for their helpful suggestions.

\bibliography{anthology,emnlp2020}
\bibliographystyle{acl_natbib}

\appendix

\section{Preliminary Experiments} \label{sec:prelim}
Here we discuss the preliminary experiments to assess the quality and computational efficiency of our posterior approximation procedure (\S\ref{sec:posterior_approx}). Recall that this approximation procedure only applies at \emph{inference}; the LM is still \emph{trained} in a typical autoregressive, left-to-right fashion.

\paragraph{Model.} Since exactly computing the RNNG's next-word distributions $t_{\bm{\phi}}(x_i | \mathbf{x}_{<i})$ involves an intractable marginalisation over all possible tree prefixes $\mathbf{y}_{<i}$, we run our experiments in the context of sequential LSTM language models, where $t_{\text{LSTM}}(x_i | \mathbf{x}_{<i})$ can be computed exactly. This setup crucially enables us to isolate the impact of approximating the posterior distribution over $x_i$ under the bidirectional context (Eq.~\ref{eq:full_posterior}) with our proposed approximation (Eq.~\ref{eq:posterior_approx_2}), without introducing further confounds stemming from the RNNG's marginal approximation procedure (Eq.~\ref{eq:marginal_approx}).

\paragraph{Dataset and preprocessing.} We train the LSTM LM on an \emph{open-vocabulary} version of the PTB,\footnote{Our open-vocabulary setup means that our results are not directly comparable to prior work on PTB language modelling \citep[\emph{inter alia}]{mikolov:2010}, which mostly employ a special ``UNK'' token for infrequent or unknown words.} in order to simulate the main experimental setup where both the RNNG teacher and BERT student are also open-vocabulary by virtue of byte-pair encoding (BPE) preprocessing. To this end, we preprocess the dataset with SentencePiece \citep{sentencepiece} BPE tokenisation, where $|\Sigma|=8,000$; we preserve all case information. We follow the empirical setup of the parsing (\S\ref{sec:tasks}) experiments, with Sections 02-21 for training, Section 22 for validation, and Section 23 for testing.

\paragraph{Model hyper-parameters.} We train the LM with 2 LSTM layers, 250 hidden units per layer, and a dropout \citep{dropout} rate of 0.2. Model parameters are optimised with stochastic gradient descent (SGD), with an initial learning rate of 0.25 that is decayed exponentially by a factor of 0.92 for every epoch after the tenth. Since our approximation relies on a separately trained right-to-left LM (Eq.~\ref{eq:posterior_approx_2}), we train this variant with the exact same hyper-parameters and dataset split as the left-to-right model. 

\paragraph{Evaluation and baselines.} We evaluate the models in terms of the average posterior negative log likelihood (NLL) and perplexity.\footnote{In practice, this perplexity is derived from simply exponentiating the average posterior negative log likelihood.} Since exact inference of the posterior is expensive, we evaluate the model only on the first 400 sentences of the test set. We compare the following variants:
\begin{itemizesquish}
\item a mixture of experts baseline that simply mixes ($\alpha=0.5$) the probabilities from the left-to-right and right-to-left LMs in an \emph{additive} fashion, as opposed to \emph{multiplicative} as in the case of our PoE-like approximation in Eq.~\ref{eq:posterior_approx_2} (\textbf{``MoE''});
\item our approximation of the posterior over $x_i$ (Eq.~\ref{eq:posterior_approx_2}), where $q(w)$ is the \emph{uniform} distribution (\textbf{``Uniform Approx.''});
\item our approximation of the posterior over $x_i$ (Eq.~\ref{eq:posterior_approx_2}), but where $q(w)$ is the \emph{unigram} distribution (\textbf{``Unigram Approx.''}); and
\item exact inference of the posterior as computed from the left-to-right model, as defined in Eq.~\ref{eq:full_posterior} (\textbf{``Exact Inference''}). 
\end{itemizesquish}

\paragraph{Discussion.} We summarise the findings in Table~\ref{tab:posterior_approx}, based on which we remark on two observations. First, the posterior NLL of our approximation procedure that makes use of the unigram distribution (\textbf{Unigram Approx.}; third row) is not much worse than that of exact inference, in exchange for a more than 50,000 times speedup\footnote{All three approximations in Table~\ref{tab:posterior_approx} have similar runtimes.} in computation time. Nevertheless, using the uniform distribution (second row) on $q(w)$ in place of the unigram one (Eq.~\ref{eq:posterior_approx_2}) results in a much worse posterior NLL. Second, combining the left-to-right and right-to-left LMs using a mixture of experts---a baseline which is not well-motivated by our theoretical analysis---yields the worst result.

\begin{table}[t]
\centering
  \resizebox{0.45\textwidth}{!}{%
\begin{tabular}{l|r|r}
\toprule
\multicolumn{1}{c|}{\textbf{Model}} & \multicolumn{1}{c|}{\textbf{Posterior NLL}} & \multicolumn{1}{c}{\textbf{Posterior Ppl.}} \\ \midrule
MoE & 3.28 & 26.58 \\ 
Uniform Approx.                     & 3.18                                   & 24.17                                   \\
Unigram Approx.                      & \textbf{2.68}                                   & \textbf{14.68}                                   \\ \hline \hline
Exact Inference                         & 2.50                                   & 12.25                       \\ \bottomrule           
\end{tabular}}
\caption{The findings from the preliminary experiments that assess the quality of our posterior approximation procedure. We compare three variants against exact inference (bottom row; Eq.~\ref{eq:full_posterior}) from the left-to-right model.}\label{tab:posterior_approx}
\end{table}

\section{RNNG Hyper-parameters} \label{sec:rnng_params}
To train the subword-augmented RNNG teacher (\S\ref{sec:rnng}), we use the following hyper-parameters that achieve the best validation perplexity from a grid search: 2-layer stack LSTMs \citep{dyer:2015} with 512 hidden units per layer, optimised by standard SGD with an initial learning rate of 0.5 that is decayed exponentially by a factor of 0.9 after the tenth epoch. We apply a dropout rate of 0.3. 
\begin{table*}[!htb]
\setlength{\tabcolsep}{4.5pt}
    \centering
    \begin{tabular}{ll|ccccccc|c}
     \toprule
      \multicolumn{2}{c|}{\multirow{2}{*}{\textbf{Model}}} & \multirow{2}{*}{\textbf{\textsc{CoLA}}} & \multirow{2}{*}{\textbf{\textsc{SST-2}}} & \multirow{2}{*}{\textbf{\textsc{MRPC}}} & \multirow{2}{*}{\textbf{\textsc{QQP}}} &
      \textbf{\textsc{MNLI}} & \multirow{2}{*}{\textbf{\textsc{QNLI}}} & \multirow{2}{*}{\textbf{\textsc{RTE}}} &
      \textbf{\textsc{GLUE}} \\
      &&&&&& \textbf{\textsc{(m/mm)}} & & & \textbf{\textsc{Avg}}\\
      \midrule
      \multirow{2}{*}{\rotatebox[origin=c]{90}{\textsc{Dev}}} & No-KD & 60.2 & 92.2 & 90.0 & 89.4 & 90.3/90.9 & 90.7 & 71.1 & 84.4  \\
      &UG-KD & 60.6 & 92.0 & 88.9  & 89.3 & 89.6/90.0 & 89.9 & 68.6 & 83.6 \\
      \midrule
      \multirow{2}{*}{\rotatebox[origin=c]{90}{\textsc{Test}}} & No-KD &  53.1 & 92.5 & 88.0 & 88.8 & 82.8/81.8 & 89.9 & 65.4 & 80.3 \\
      &UG-KD & 55.3 & 91.2 & 87.6 & 88.7 & 81.9/80.8 & 89.5 & 65.0 & 80.0 \\
      \bottomrule
    \end{tabular}
    \caption{Summary of the full results on GLUE, comparing the \textbf{No-KD} baseline with the \textbf{UG-KD} structure-distilled BERT (\S\ref{sec:experimental_setup}). We select the 1-best fine-tuning hyper-parameter (including random seed) on the validation set, which we then evaluate on the test set.} \label{tbl:results1}
    \vspace{-0.5cm}
\end{table*}
\section{Right-to-left RNNG} \label{sec:right-to-left}
Here we illustrate the oracle action sequences that we use to train the right-to-left RNNG teacher as part of our approximation of the posterior distribution over $x_i$ (Eq.~\ref{eq:posterior_approx_2}). Recall that the standard RNNG incrementally builds the phrase-structure tree through a top-down, left-to-right traversal in a \emph{depth-first} fashion. Hence, the right-to-left RNNG employs a similar top-down, depth-first traversal strategy, although the children of each node are recursively expanded in a right-to-left fashion. 

We provide example action sequences (Table~\ref{tab:action_sequences}) for the subword-augmented left-to-right and right-to-left RNNGs, respectively, for an example ``(S (NP (WORD \emph{The}) (WORD \emph{d \#\#og})) (VP (WORD \emph{ba \#\#rk \#\#s})))'', where tokens prefixed by ``\emph{\#\#}'' are subword units.

\begin{table*}[t]
\centering 
\resizebox{\textwidth}{!}{%
\begin{tabular}{r|l|l} \toprule
\multicolumn{1}{c|}{\textbf{Step}} & \multicolumn{1}{c|}{\textbf{Stack Content}}                               & \multicolumn{1}{c}{\textbf{Action}} \\ \midrule
\multicolumn{3}{c}{\textbf{Left-to-right RNNG}}                                                                                                             \\ \hline \hline
0                                 &                                                                          & NT(S)                               \\
1                                 & (S                                                                       & NT(NP)                              \\
2                                 & (S $\vert$ (NP                                                                 & NT(WORD)                            \\
3                                 & (S $\vert$ (NP $\vert$ (WORD                                                         & GEN(\emph{The})                            \\
4                                 & (S $\vert$ (NP $\vert$ (WORD $\vert$ \emph{The}                                                   & REDUCE                              \\
5                                 & (S $\vert$ (NP $\vert$ (WORD \emph{The})                                                    & NT(WORD)                            \\
6                                 & (S $\vert$ (NP $\vert$ (WORD \emph{The}) $\vert$ (WORD                                            & GEN(\emph{d})                              \\
7                                 & (S $\vert$ (NP $\vert$ (WORD \emph{The}) $\vert$ (WORD $\vert$ \emph{d}                                        & GEN(\emph{\#\#og})                         \\
8                                 & (S $\vert$ (NP $\vert$ (WORD \emph{The}) $\vert$ (WORD $\vert$ \emph{d} $\vert$ \emph{\#\#og}                               & REDUCE                              \\
9                                 & (S $\vert$ (NP $\vert$ (WORD \emph{The}) $\vert$ (WORD \emph{d \#\#og})                                  & REDUCE                              \\
10                                & (S $\vert$ (NP (WORD \emph{The}) (WORD \emph{d \#\#og}))                                     & NT(VP)                              \\
11                                & (S $\vert$ (NP (WORD \emph{The}) (WORD \emph{d \#\#og})) $\vert$ (VP                               & NT(WORD)                            \\
12                                & (S $\vert$ (NP (WORD \emph{The}) (WORD \emph{d \#\#og})) $\vert$ (VP $\vert$ (WORD                       & GEN(\emph{ba})                             \\
13                                & (S $\vert$ (NP (WORD \emph{The}) (WORD \emph{d \#\#og})) $\vert$ (VP $\vert$ (WORD $\vert$ \emph{ba}                  & GEN(\emph{\#\#rk})                         \\
14                                & (S $\vert$ (NP (WORD \emph{The}) (WORD \emph{d \#\#og})) $\vert$ (VP $\vert$ (WORD $\vert$ \emph{ba} $\vert$ \emph{\#\#rk}         & GEN(\emph{\#\#s})                          \\
15                                & (S $\vert$ (NP (WORD \emph{The}) (WORD \emph{d \#\#og})) $\vert$ (VP $\vert$ (WORD $\vert$ \emph{ba} $\vert$ \emph{\#\#rk} $\vert$ \emph{\#\#s} & REDUCE                              \\
16                                & (S $\vert$ (NP (WORD \emph{The}) (WORD \emph{d \#\#og})) $\vert$ (VP $\vert$ (WORD \emph{ba \#\#rk \#\#s})      & REDUCE                              \\
17                                & (S $\vert$ (NP (WORD \emph{The}) (WORD \emph{d \#\#og})) $\vert$ (VP (WORD \emph{ba \#\#rk \#\#s}))       & REDUCE                              \\
18                                & (S (NP (WORD \emph{The}) (WORD \emph{d \#\#og})) (VP (WORD \emph{ba \#\#rk \#\#s})))          &   \\ \midrule
\multicolumn{3}{c}{\textbf{Right-to-left RNNG}} \\ \hline \hline
0                                 &                                                                          & NT(S)                               \\
1                                 & (S                                                                       & NT(VP)                              \\
2                                 & (S $\vert$ (VP                                                                & NT(WORD)                            \\
3                                 & (S $\vert$ (VP $\vert$ (WORD                                                         & GEN(\emph{\#\#s})                            \\
4                                 & (S $\vert$ (VP $\vert$ (WORD $\vert$ \emph{\#\#s}                                                   & GEN(\emph{\#\#rk})                             \\
5                                 & (S $\vert$ (VP $\vert$ (WORD $\vert$ \emph{\#\#s} $\vert$ \emph{\#\#rk}                                                  & GEN(\emph{ba})                            \\
6                                 & (S $\vert$ (VP $\vert$ (WORD $\vert$ \emph{\#\#s} $\vert$ \emph{\#\#rk}  $\vert$ \emph{ba}                                           & REDUCE                             \\
7                                 & (S $\vert$ (VP $\vert$ (WORD \emph{\#\#s \#\#rk ba})                                        & REDUCE                          \\
8                                 & (S $\vert$ (VP (WORD \emph{\#\#s \#\#rk ba}))                              & NT(NP)                              \\
9                                 & (S $\vert$ (VP (WORD \emph{\#\#s \#\#rk ba})) $\vert$ (NP                                  & NT(WORD)                              \\
10                                & (S $\vert$ (VP (WORD \emph{\#\#s \#\#rk ba})) $\vert$ (NP $\vert$ (WORD                                     & GEN(\emph{\#\#og})                             \\
11                                & (S $\vert$ (VP (WORD \emph{\#\#s \#\#rk ba})) $\vert$ (NP $\vert$ (WORD $\vert$ \emph{\#\#og}                               & GEN(\emph{d})                            \\
12                                & (S $\vert$ (VP (WORD \emph{\#\#s \#\#rk ba})) $\vert$ (NP $\vert$ (WORD $\vert$ \emph{\#\#og} $\vert$ \emph{d}                       & REDUCE                            \\
13                                & (S $\vert$ (VP (WORD \emph{\#\#s \#\#rk ba})) $\vert$ (NP $\vert$ (WORD \emph{\#\#og d})                  & NT(WORD)                         \\
14                                & (S $\vert$ (VP (WORD \emph{\#\#s \#\#rk ba})) $\vert$ (NP $\vert$ (WORD \emph{\#\#og d}) $\vert$ (WORD         & GEN(\emph{The})                          \\
15                                & (S $\vert$ (VP (WORD \emph{\#\#s \#\#rk ba})) $\vert$ (NP $\vert$ (WORD \emph{\#\#og d}) $\vert$ (WORD $\vert$ \emph{The} & REDUCE                              \\
16                                & (S $\vert$ (VP (WORD \emph{\#\#s \#\#rk ba})) $\vert$ (NP $\vert$ (WORD \emph{\#\#og d}) $\vert$ (WORD \emph{The})      & REDUCE                              \\
17                                & (S $\vert$ (VP (WORD \emph{\#\#s \#\#rk ba})) $\vert$ (NP (WORD \emph{\#\#og d}) (WORD \emph{The}))       & REDUCE                              \\
18                                & (S (VP (WORD \emph{\#\#s \#\#rk ba})) (NP (WORD \emph{\#\#og d}) (WORD \emph{The})))          &   \\ 
\bottomrule                                 
\end{tabular}}
\caption{Sample stack contents and the corresponding gold action sequences for a simple example ``(S (NP (WORD \emph{The}) (WORD \emph{d \#\#og})) (VP (WORD \emph{ba \#\#rk \#\#s})))'', under both the left-to-right and right-to-left subword-augmented RNNGs (\S\ref{sec:rnng}). The symbol ``$\vert$'' denotes separate entries on the stack. At the end of the generation process, the stack contains one composite embedding that represents the entire tree.}\label{tab:action_sequences}
\end{table*}

\section{BERT Hyper-parameters} \label{sec:bert_params}
Here we outline the hyper-parameters of the BERT student in terms of pretraining data creation, masked LM pretraining, and GLUE fine-tuning. 

\paragraph{Pretraining data creation.} We use the same codebase\footnote{\url{https://github.com/google-research/bert}.} and pretraining data as \citet{devlin_2019}, which are derived from a mixture of Wikipedia and Books text corpora. To train our structure-distilled BERTs, we sample a masking from these corpora following the same hyper-parameters used to train the original BERT$_{\text{BASE}}$-Cased model: a maximum sequence length of 512, a per-word masking probability of 0.15 (up to a maximum of 76 masked tokens in a 512-length sequence), a dupe factor of 10. We apply a random seed of 12345. We preprocess the raw dataset using NLTK tokenisers, and then apply the same (BPE-augmented) vocabulary and WordPiece tokenisation as in the original BERT model. All other hyper-parameters are set to the same values as in the publicly released original BERT model.

\paragraph{Masked LM pretraining.} We train all model variants (including the no distillation/standard BERT baseline for fair comparison) with the following hyper-parameters: a batch size of 256 sequences and an initial Adam learning rate of $3e^{-4}$ (as opposed to $1e-4$ in the original BERT model). Following \citet{devlin_2019}, we pretrain our models for 1M steps. All other hyper-parameters are similarly set to their default values.

\paragraph{GLUE fine-tuning.} For each GLUE task, we fine-tune the BERT model by running a grid search over five potential learning rates $\{5e^{-6}, 1e^{-5}, 2e^{-5}, 3e^{-5}, 5e^{-5}\}$, two potential batch sizes $\{16, 32\}$, and five random seeds, in order to better account for variance. This setup leads to 50 fine-tuning configurations for each GLUE task. Following \citet{devlin_2019}, we train each fine-tuning configuration for 4 epochs. 

\paragraph{Structured prediction fine-tuning.} For each structured prediction model, we use the model's default BERT fine-tuning settings for learning rate, batch size, and learning rate warmup schedule. We use the default settings for BERT$_{\text{BASE}}$, if these are available, and the default settings for BERT$_{\text{LARGE}}$ otherwise. These settings are:
\begin{itemize}
\item In-order phrase-structure parser: a BERT learning rate of $2e^{-5}$, a batch size of 32, and a warmup period of 160 updates.
\item HPSG dependency parser: a BERT learning rate of $5e^{-5}$, a batch size of 150, and a warmup period of 160 updates.
\item Coreference resolution model: a BERT learning rate of $1e^{-5}$, a batch size of 1 document, and a warmup period of 2 epochs.
\item Semantic role labelling model: a BERT learning rate of  $5e^{-5}$ and a batch size of 32.
\end{itemize}

\paragraph{Comparison.} Our no distillation baseline differs from the publicly released BERT$_{\text{BASE}}$-Cased model in its larger pretraining learning rate ($3e^{-4}$ as opposed to $1e^{-4}$) that we empirically find to work better on most of the tasks. Overall, our no distillation baseline slightly outperforms the publicly released model on all the structured prediction tasks except coreference resolution, where it performs slightly worse. Furthermore, our no distillation baseline also performs slightly better than the official pretrained BERT$_{\text{BASE}}$ on most of the GLUE tasks, although the difference in aggregate GLUE performance is fairly minimal ($<0.5$).

\section{Full GLUE Results} \label{sec:full_glue}
We summarise the full GLUE results for the \textbf{No-KD} baseline and the \textbf{UG-KD} structure-distilled BERT in Table~\ref{tbl:results1}.

\end{document}